\documentclass[letterpaper, 10 pt, journal, twoside]{IEEEtran}

\usepackage{iftex}
\ifXeTeX
\usepackage{fontspec}
\fi
\usepackage{graphics}
\usepackage{graphicx}
\usepackage{amsmath}
\usepackage{amssymb}
\usepackage{amsfonts}
\usepackage{cite}
\usepackage{xcolor}
\usepackage{booktabs}
\usepackage{colortbl}
\usepackage{algorithm}
\usepackage{algorithmic}

\makeatletter
\g@addto@macro\fs@ruled{%
  \def\@fs@pre{\begingroup\normalcolor\hrule height.8pt depth0pt \kern2pt}%
  \def\@fs@post{\kern2pt\hrule\relax\endgroup}}
\makeatother
\usepackage{bm}
\usepackage{makecell}

\newcommand{\R}{\mathbb{R}}
\newcommand{\mW}{\bm{W}}
\newcommand{\mU}{\bm{U}}
\newcommand{\mV}{\bm{V}}
\newcommand{\mI}{\bm{I}}
\newcommand{\mH}{\bm{H}}
\newcommand{\mP}{\bm{P}}

\newcommand{\mF}{\bm{F}}
\newcommand{\vx}{\bm{x}}
\newcommand{\vy}{\bm{y}}
\newcommand{\vz}{\bm{z}}

\newcommand{\vr}{\bm{r}}

\newcommand{\vu}{\bm{u}}
\newcommand{\vv}{\bm{v}}
\newcommand{\vp}{\bm{p}}

\newcommand{\ve}{\bm{e}}
\newcommand{\veps}{\bm{\epsilon}}
\newcommand{\va}{\bm{a}}
\newcommand{\vo}{\bm{o}}
\newcommand{\vs}{\bm{s}}
\newcommand{\vc}{\bm{c}}

\newcommand{\method}{SR-DP}
\newcommand{\best}[1]{\textbf{#1}}      % best entry per column: plain bold

\title{Spectrally Routed Diffusion Policy for Multi-Task Robotic Manipulation}

\author{
Mingcong Li$^{1,4,*}$,
Yihan Zhou$^{2,*}$,
Jiayu Qu$^{1,4}$,
Xueyang Guo$^{2}$,
and Yilin Mo$^{3,\dagger}$%
\thanks{This work was supported by the National Natural Science Foundation of China under Grants 62461160313, 62273196, 62192752 and the BNRist project (No. BNR2024TD03003).}%
\thanks{$^{*}$Mingcong Li and Yihan Zhou contributed equally to this work.}%
\thanks{$^{\dagger}$Corresponding author: Yilin Mo (e-mail: {\tt\footnotesize ylmo@tsinghua.edu.cn}).}%
\thanks{$^{1}$Mingcong Li and Jiayu Qu are with the School of Automation, Beijing Institute of Technology, Beijing 100081, China.}%
\thanks{$^{2}$Yihan Zhou and Xueyang Guo are with the Department of Automation, Tsinghua University, Beijing 100084, China, and also with the Beijing National Research Center for Information Science and
Technology (BNRist), Tsinghua University, Beijing 100084, China.}%
\thanks{$^{3}$Yilin Mo is with the Department of Automation, Tsinghua University, Beijing 100084, China, and also with the Beijing Key Laboratory of Embodied Intelligence Systems, Beijing, China, and
also with the Institute for Embodied Intelligence and Robotics, Tsinghua University, Beijing 100084, China.}%
\thanks{$^{4}$LingYu Robotics, Beijing 100086, China.}%
}

\begin{document}
\bstctlcite{IEEEexample:BSTcontrol}

\maketitle
\thispagestyle{empty}
\pagestyle{empty}

% ============================================================
%  Body  (one file per section, see sections/)
% ============================================================
%========================================================================
\begin{abstract}
    Diffusion-based policies have achieved strong performance on single-task robotic manipulation.
    However, a single model trained jointly on multiple tasks consistently underperforms the per-task specialists trained on the same demonstrations.
    We focus on two challenges.
    First, the convolutional encoders commonly used in diffusion-based policies produce visual conditions that are not discriminative enough.
    Second, gradients from incompatible action modes update the shared denoiser's dense parameters along the same directions, causing destructive interference during joint training.
    To address both, we propose \method{} (\textbf{S}pectrally \textbf{R}outed \textbf{D}iffusion \textbf{P}olicy).
    A frozen VGGT encoder with a spatial-compression adapter produces compact multi-scale semantic-geometric tokens, a task-discriminative visual condition for tasks that share a workspace.
    We first train the denoiser as a standard diffusion policy, then decompose selected linear layers by SVD into a frozen basis of orthogonal rank-one experts, adapted through spectral routing and Householder input rotations.
    On six RoboTwin tasks, \method{} averages a $68.4\%$ success rate, surpassing multi-task VO-DP by $21.8$ points and the strongest per-task baseline by $3.8$ points.
    On four real-world bimanual tasks, Stage~2 reaches a $43.0\%$ average success rate and improves over Stage~1 by $19.0$ points.
\end{abstract}

\begin{IEEEkeywords}
Imitation Learning, Deep Learning in Grasping and Manipulation, Bimanual Manipulation.
\end{IEEEkeywords}

%========================================================================
\section{Introduction}
\label{sec:intro}

\IEEEPARstart{D}{iffusion} Policy~\cite{chi2024diffusionpolicy} has emerged as a dominant paradigm for visuomotor imitation learning, achieving strong results on single-task benchmarks.
However, this success does not extend to the multi-task setting.
Joint training across diverse tasks challenges a shared policy's ability to model task-specific action distributions~\cite{liu2025fdp,wang2024sparse_dp_moe}.
Enlarging the backbone or collecting more demonstrations only partially closes this gap, while the underlying cause remains poorly understood.

We attribute this degradation to two structural limitations.
On the perception side, the convolutional encoders commonly used in diffusion-based policies~\cite{chi2024diffusionpolicy} apply global pooling that compresses each observation into a single vector, discarding the spatial layout that distinguishes tasks within a shared workspace.
Conversely, vision-language-action (VLA) methods preserve this detail but rely on large vision-language backbones with full token attention~\cite{black2024pi0,kim_openvla_2024}.
Compact conditions that keep multi-scale spatial structure remain uncommon.
On the action side, different tasks induce distinct conditional action distributions.
Their gradients update the same weight directions of the shared denoiser in conflicting ways~\cite{yu2020gradient,liu2021conflict}, causing joint training to converge to a compromise that fits no single task.
Most existing approaches distribute capacity across multiple subnetworks via Mixture-of-Experts or action-distribution decomposition~\cite{wang2024sparse_dp_moe,reuss2024mode,liu2025fdp}.
However, these modules are themselves newly added, unconstrained dense parameters, leaving capacity duplication and inter-task gradient interference unresolved.

Rather than introducing separate dense modules for different tasks, we constrain task adaptation to the structure encoded by the shared denoiser.
For a linear weight $\mW$, the SVD $\mW=\mU\,\mathrm{diag}(\bm{\sigma})\,\mV^\top=\sum_{d=1}^{D}\sigma_d\vu_d\vv_d^\top$ expresses the layer as a set of orthogonal rank-one components.
We freeze the base factors $\mU$, $\mV$, and $\bm{\sigma}$; a compact router modulates their effective spectral coefficients $\hat{\bm{\sigma}}$, while Householder reflections rotate the input directions under an orthogonality constraint.
Because the output directions $\vu_d$ remain fixed and orthogonal, changes in $\hat{\sigma}_d$ act along separated output directions before nonlinear mixing, constraining task-specific adaptation without unrestricted dense-weight updates.

Building on this idea, we propose \method{} (Fig.~\ref{fig:overview}), a multi-task diffusion policy that pairs a visual encoder with a spectrally routed denoiser.
The visual encoder uses a frozen VGGT backbone~\cite{wang2025vggt}, whose alternating frame-wise and global self-attention blocks capture both object identity and spatial arrangement that distinguish tasks sharing a workspace.
A spatial-compression adapter condenses VGGT's dense per-view tokens into a compact multi-scale condition for the denoiser, while an auxiliary task-prediction head encourages the resulting condition to stay task-discriminative.
The denoiser is an AdaLN-Zero DiT~\cite{peebles2023dit} trained in two stages.
Stage~1 jointly trains the denoiser without task conditioning, yielding a shared base policy.
Stage~2 decomposes selected linear layers as above and adds a Householder rotation and a router conditioned on task identity, token state, and diffusion step.
This conversion preserves the base policy's learned mapping at initialization.

We evaluate \method{} as a multi-task model on six RoboTwin tasks~\cite{mu2024robotwin} and on four real-world tasks collected on a teleoperated bimanual platform.
A single \method{} model reaches $68.4\%$ average success on the six jointly trained RoboTwin tasks, exceeding the per-task VO-DP baseline by $3.8$ points and the multi-task VO-DP baseline by $21.8$; on our four real-world tasks designed with similar scenes but distinct action distributions, the Stage~2 model achieves $43.0\%$ average success, improving over Stage~1 by $19.0$ points.

\textit{Contributions:} This work advances multi-task diffusion policies through: (1) A spectral-routing adaptation for diffusion policies that combines Householder input rotations with routing conditioned on task identity, token state, and diffusion step, keeping the base SVD factors frozen while limiting inter-task interference and training fewer than $5\%$ of policy-head parameters in Stage~2; (2) A compact, task-discriminative visual condition built from a frozen VGGT encoder, a spatial-compression adapter, and a task-prediction head, enabling a DP-scale denoiser to serve multiple tasks that share a workspace; and (3) Experiments on RoboTwin and a real bimanual platform showing that a single multi-task model matches or exceeds the per-task baselines, with an analysis associating the gain with less opposition among spectral-coordinate loss sensitivities.

%========================================================================
\section{Related Work}
\label{sec:related}

\subsection{Diffusion Policies for Multi-Task Manipulation}
Diffusion Policy~\cite{chi2024diffusionpolicy} casts the policy as a conditional denoising process and achieves strong single-task imitation results.
Follow-up work swaps the image input for a point cloud (DP3~\cite{ze20243ddp}) or scales up the denoiser (RDT-1B~\cite{liu2025rdtb}), but per-task accuracy still drops once a single model is trained jointly across many tasks.

To distribute capacity across tasks, several diffusion policies add dedicated submodules: SDP~\cite{wang2024sparse_dp_moe} inserts sparse Mixture-of-Experts layers with a mutual-information objective in place of the usual load-balancing loss, MoDE~\cite{reuss2024mode} routes by denoising step so capacity specializes per noise level, and FDP~\cite{liu2025fdp} factorizes the action distribution into composable components combined at inference.
Complementary approaches target interference through optimization or representation structure: PCGrad and CAGrad modify shared updates to reduce gradient conflict~\cite{yu2020gradient,liu2021conflict}, whereas SMP learns an orthogonal skill basis with sticky expert routing for multi-task diffusion policies~\cite{hao2026smp}.
Large VLA models~\cite{kim_openvla_2024,black2024pi0} instead improve broad multi-task capacity by scaling data and compute.
\method{} instead constrains adaptation through routed spectral coefficients and Householder input rotations, while keeping the base SVD factors frozen.

\subsection{Vision Encoders for Manipulation}
Pretrained vision models are increasingly used as observation encoders for manipulation.
Early imitation policies such as DP~\cite{chi2024diffusionpolicy} and ACT~\cite{zhao2023act} rely on general-purpose backbones like ResNet, which provide limited fine-grained features.
Recent visual foundation models offer richer ones: DINOv2~\cite{oquab2024dinov2} produces dense semantic features, and VGGT~\cite{wang2025vggt} further exposes cross-view geometric features on top of them.
Several works adopt VGGT as a frozen encoder for manipulation: VGGT-DP~\cite{he2025vggtdp} integrates it into a single-task diffusion policy with proprio-guided learning, and VO-DP~\cite{li2025vodp} fuses its semantic and geometric features for vision-only diffusion policies.
\method{} targets the multi-task setting, where the visual condition must stay task-discriminative across tasks sharing a workspace while remaining compact enough for a DP-scale denoiser.

\subsection{Orthogonal and Low-Rank Adaptation}
Parameter-efficient fine-tuning preserves pretrained structure by constraining how weights are updated.
OFT~\cite{qiu2023controlling} constrains updates to orthogonal transformations, BOFT~\cite{liu2024parameter} accelerates this via butterfly factorization, and HRA~\cite{yuan2024bridging} applies Householder reflections to stay within the original column space.
SVDiff fine-tunes static singular values for compact diffusion adaptation~\cite{han2023svdiff}.
MoORE~\cite{xu2025moore}, proposed for multi-task language-model adaptation, decomposes a pretrained linear layer by SVD and treats the rank-one terms as experts whose scaling factors are routed.
\method{} adopts the same spectral-routing principle but brings it to a pretrained diffusion denoiser: it applies the token-state branch independently to denoiser tokens, augments it with a diffusion-step bias, selects which layers to route by their condition number rather than decomposing every layer, and converts an already-converged diffusion policy in a second stage so the routed network initially reproduces it.

\begin{figure*}[t]
\centering
\includegraphics[width=0.74\textwidth]{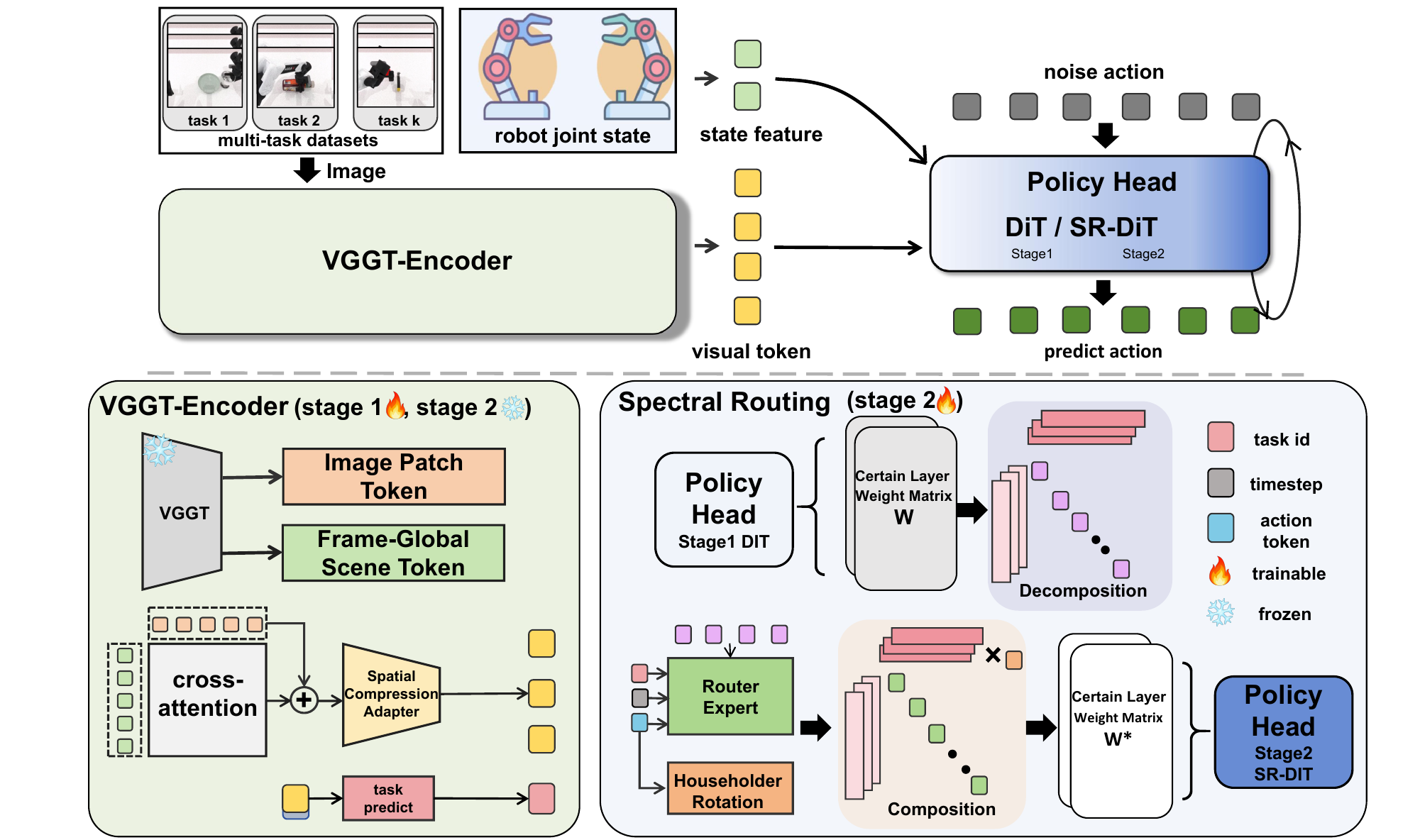}%
\vspace{-4pt}
\caption{\textbf{Overview of \method{}.} A frozen VGGT encoder with a spatial-compression adapter produces $14$ tokens per view, which condition an AdaLN-Zero DiT denoiser. Selected linear layers of the denoiser are decomposed via SVD into rank-one experts and modulated by a compact router (Stage~2).}
\label{fig:overview}
\vspace{-10pt}
\end{figure*}

%========================================================================
\section{Method}
\label{sec:method}

We consider imitation learning over $K$ manipulation tasks, indexed by $k\in\{1,\dots,K\}$, each providing a demonstration set $\mathcal{D}_k = \{(\vo_t, \va_t)\}$ of observation--action pairs collected for task $k$.
An observation $\vo_t = (\{\mI_t^m\}_{m=1}^M, \vs_t)$ consists of $M$ RGB images and a proprioceptive state $\vs_t \in \R^{D_s}$, and an action $\va_t \in \R^{H \times D_a}$ is a chunk of $H$ future actions, each a $D_a$-dimensional joint command.
Following~\cite{chi2024diffusionpolicy}, we model the policy as a conditional DDPM~\cite{ho2020ddpm}: at diffusion step $n$ the clean action $\va_t$ is corrupted into $\va_t^n$, and a denoiser $\veps_\theta$ predicts the injected noise $\veps$,
\begin{equation}
\label{eq:ddpm}
    \mathcal{L}_{\text{DM}} = \mathbb{E}_{n,\veps,(\vo_t,\va_t)\sim\bigcup_k\mathcal{D}_k} \big\| \veps - \veps_\theta(\va_t^n, n, \vc_t) \big\|_2^2
\end{equation}
where $\vc_t = \mathrm{Enc}(\vo_t)$ is the observation conditioning from the encoder ({\S}\ref{sec:encoder}), comprising the visual tokens $\vc^{\text{vis}}_t$ and a proprioceptive embedding of $\vs_t$.

In \method{} (Fig.~\ref{fig:overview}), a frozen VGGT backbone~\cite{wang2025vggt} and a lightweight spatial-compression adapter map multi-view images in $\vo_t$ to the compact visual condition $\vc^{\text{vis}}_t$ for cross-attention ({\S}\ref{sec:encoder}).
Conditioned on $\vc_t$, the policy head is an AdaLN-Zero DiT~\cite{peebles2023dit} denoiser trained in two stages ({\S}\ref{sec:training}, Fig.~\ref{fig:two_stage}).
Stage~1 trains it jointly across all tasks as a standard diffusion policy, with no task-specific parameters.
Stage~2 freezes the base policy, decomposes selected linear layers by SVD into a frozen orthogonal basis of rank-one experts, and trains a compact router conditioned on task identity, denoiser token state, and diffusion step ({\S}\ref{sec:moore}), constraining task adaptation through spectral coordinates without rewriting the shared denoiser weights.

\begin{figure}[t]
\centering
\includegraphics[width=\columnwidth]{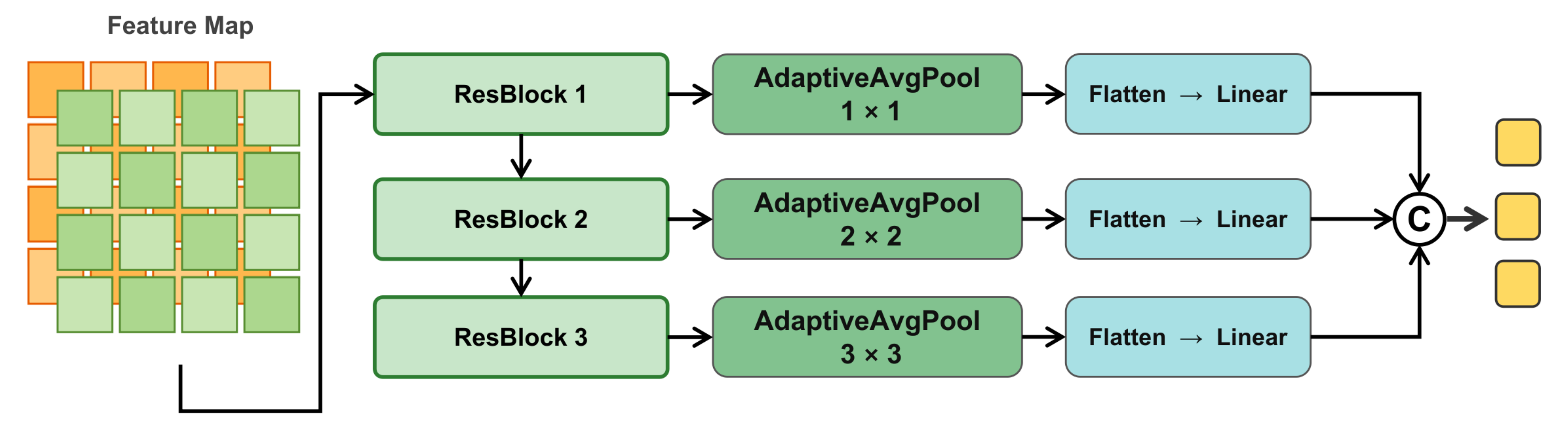}%
\vspace{-6pt}
\caption{\textbf{Spatial-compression adapter.} Three residual blocks form a coarse-to-fine pyramid; each level is pooled at $1{\times}1$, $2{\times}2$ or $3{\times}3$, flattened, linearly projected, and concatenated into $14$ tokens per view (Eq.~\ref{eq:multiscale_pool}).}
\label{fig:satc}
\vspace{-10pt}
\end{figure}

%------------------------------------------------------------------------
\subsection{VGGT Encoder}
\label{sec:encoder}

The encoder is required to produce a visual condition that is task-discriminative enough for a single policy to disambiguate multiple tasks from RGB observations alone, and compact enough for the downstream denoiser to consume directly.
We use a frozen VGGT-1B backbone~\cite{wang2025vggt}, pretrained on multi-view 3D reconstruction, with alternating frame-wise and global self-attention blocks.
At an intermediate block $\ell^\star$, VGGT exposes two token streams per view: a frame stream $\mF^{\text{frame}}_{\ell^\star}\in\R^{M\times P\times C}$ that retains per-view geometry, and a global stream $\mF^{\text{global}}_{\ell^\star}\in\R^{M\times P\times C}$ that carries cross-view semantic context.
VGGT also provides image patch tokens $\bm{Z}^{\text{img},m}\in\R^{P\times C}$.
For each view, we concatenate the frame-wise and global streams into frame-global scene tokens and fuse them with the image patch tokens through a lightweight cross-attention block,
\begin{equation}
\label{eq:vggt_fusion}
    \mF^{\text{scene},m} =
    \mathrm{CrossAttn}\!\left(
    \phi([\mF^{\text{frame},m}_{\ell^\star}\,\|\,\mF^{\text{global},m}_{\ell^\star}]),
    \bm{Z}^{\text{img},m}
    \right)
\end{equation}
where $\phi$ is a channel projection, the first argument supplies queries, and $\bm{Z}^{\text{img},m}$ supplies keys and values.
This yields a fused feature map that combines VGGT's frame-global scene representation with fine-grained image patch detail.

Feeding $\mF^{\text{scene}}$ directly to the denoiser is prohibitive, since $P$ is several hundred per view, while a single global-pooled vector would discard most of the dense detail VGGT produces.
For each view, we first reshape the $P$ patch tokens back to their VGGT spatial grid using the patch positions, obtaining $\mF^{m,0}_{\mathrm{ad}}\in\R^{H'\times W'\times C}$.
The residual adapter then forms a three-stage pyramid, and each stage is pooled at a different spatial resolution (Fig.~\ref{fig:satc}),
\begin{equation}
\label{eq:multiscale_pool}
\begin{aligned}
    \mF^{m,s}_{\mathrm{ad}} &=
    \mathrm{ResBlock}_{s}(\mF^{m,s-1}_{\mathrm{ad}}),\\
    \mathcal{G}^{m,s} &=
    \mathrm{Pool}_{r_s\times r_s}(\mF^{m,s}_{\mathrm{ad}})
    = \{\vp^{m,s}_j\}_{j=1}^{r_s^2},\\
    \vc^{\text{vis},m} &=
    \big[\,\Pi_s(\vp^{m,s}_j)\,\big]_{s=1,\dots,3;\ j=1,\dots,r_s^2}
\end{aligned}
\end{equation}
Here $s=1,2,3$, $(r_1,r_2,r_3)=(1,2,3)$, $\mathcal{G}^{m,s}$ is the $r_s{\times}r_s$ pooled grid at stage $s$, $\vp^{m,s}_j\in\R^{C}$ is its $j$-th cell feature, $\Pi_s$ is a stage-specific linear projection mapping each cell to one token, and $[\cdot]$ denotes concatenation over all stages and cells.
This yields $1+4+9=14$ tokens per view that summarize the scene at whole-scene, quadrant, and finer resolutions.
A learnable view embedding is added, and the per-view tokens $\{\vc^{\text{vis},m}\}_{m=1}^M$ are concatenated into the visual condition $\vc^{\text{vis}}_t$, which the denoiser consumes as cross-attention keys and values.
A lightweight task head mean-pools the concatenated spatial-token set $\mathcal{T}_t$ and predicts the task identifier,
\begin{equation}
    p(k\mid\vo_t) =
    \mathrm{softmax}\bigg(\mathrm{MLP}\Big(\frac{1}{|\mathcal{T}_t|}\sum_{\vz\in\mathcal{T}_t}\vz\Big)\bigg)
\end{equation}
and is trained jointly with the denoiser under a cross-entropy objective.
The auxiliary task loss encourages visual tokens to encode task identity; at inference, the frozen TaskHead infers visually identifiable task intent for the Stage~2 router.

\begin{figure}[t]
\centering
\includegraphics[width=0.76\columnwidth]{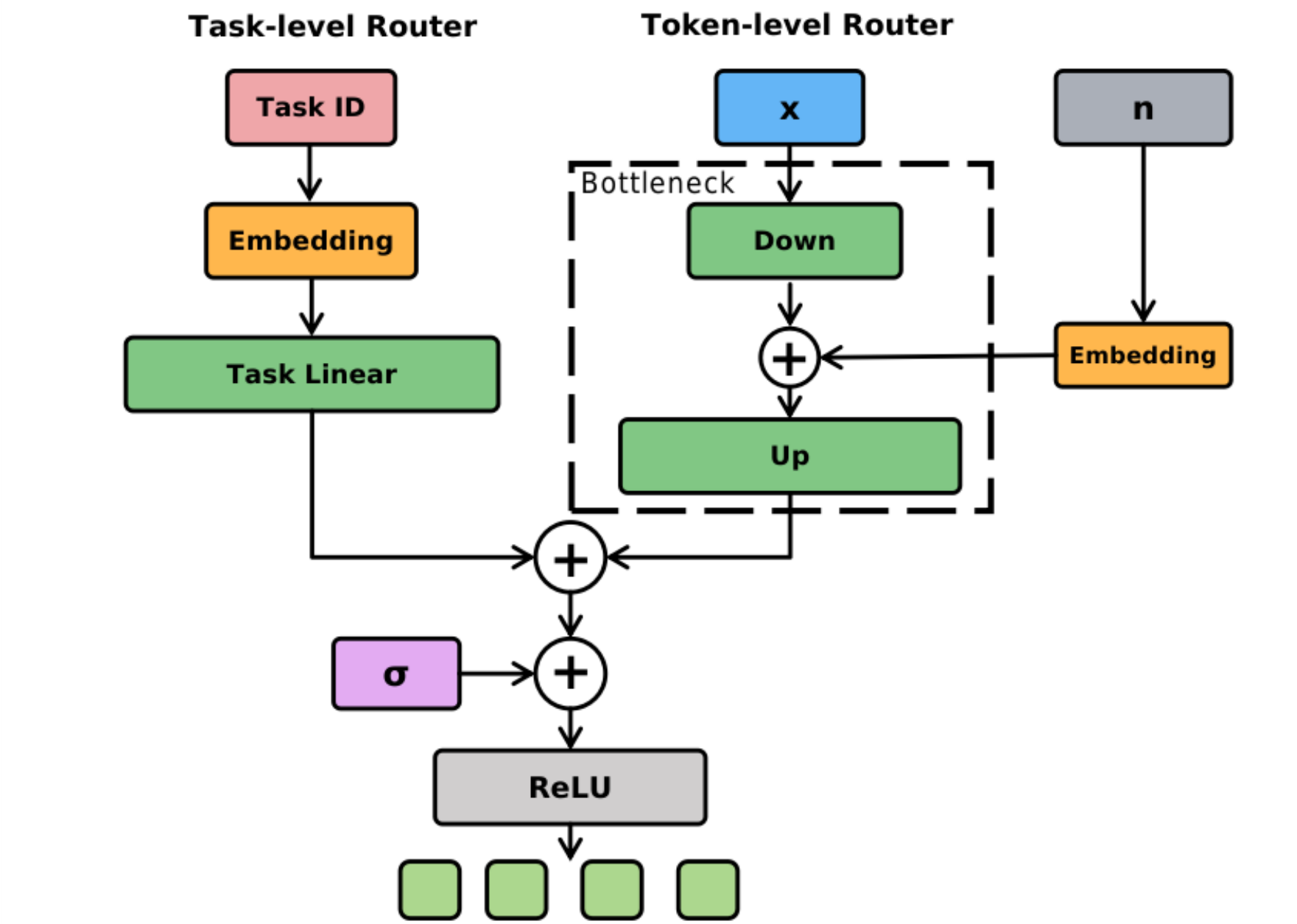}%
\vspace{-4pt}
\caption{\textbf{Router} that produces token-wise $\hat{\bm{\sigma}}_{b,q}$ in Eq.~\ref{eq:effective_sigma}. A task-level branch ($\mP\ve_{k_b}$) is shared across denoiser tokens, while a token-state bottleneck takes $\vx_{b,q}$ and adds the diffusion-step bias $\mW_n\mathrm{emb}(n_b)$.}
\label{fig:moore_layer}
\vspace{-10pt}
\end{figure}

%------------------------------------------------------------------------
\subsection{Spectrally Routed Policy Head}
\label{sec:moore}

The denoiser is an AdaLN-Zero DiT~\cite{peebles2023dit} of $L$ transformer blocks.
Each block carries a self-attention, a cross-attention on the visual condition $\vc^{\text{vis}}_t$, and an MLP, all modulated by an AdaLN layer driven by the diffusion step $n$ and the proprioceptive embedding of $\vs_t$.
After joint training, its well-conditioned linear layers are replaced by an orthogonal rank-one expert form, and a compact router (Fig.~\ref{fig:moore_layer}) drives these experts through three sources of conditioning.

For a trained weight $\mW \in \R^{D_o \times D_i}$ with bias $\bm{b}$, the thin SVD gives
\begin{equation}
\label{eq:svd}
    \mW = \mU\,\mathrm{diag}(\bm{\sigma})\,\mV^\top = \sum_{d=1}^{D} \sigma_d\,\vu_d\vv_d^\top
\end{equation}
with $D = \min(D_o,D_i)$ and orthonormal columns $\vu_d,\vv_d$.
Following MoORE~\cite{xu2025moore}, we freeze $\mU$, $\mV$, and $\bm{\sigma}$, and treat each $\vu_d\vv_d^\top$ as a rank-one expert gated by the scalar activation $\sigma_d$.
A perturbation $\delta_d$ applied to a single $\sigma_d$ changes the layer output only along the direction $\vu_d$,
\[
    (\mW + \delta_d \vu_d\vv_d^\top)\vx = \mW\vx + \delta_d\,\vu_d\,\langle \vv_d, \vx \rangle
\]
Within one routed layer, perturbations to different $\sigma_d$ therefore act through orthogonal output directions before subsequent nonlinear mixing.

For each sample $b$ and denoiser token $q$, the router produces an additive shift on the singular values,
\begin{equation}
\label{eq:effective_sigma}
    \hat{\bm{\sigma}}_{b,q} =
    \mathrm{ReLU}\big(\bm{\sigma} + \mP\ve_{k_b}
    + \mathrm{up}(\mathrm{down}(\vx_{b,q})
    + \mW_n\,\mathrm{emb}(n_b))\big)
\end{equation}
where $\vx_{b,q}$ is the hidden state of the $q$-th action token in sample $b$.
The task term $\mP\ve_{k_b}$ supplies a coarse per-task offset and is broadcast to all tokens of the sample, while the hidden-state bottleneck is applied independently to each token.
The additive $\mathrm{emb}(n_b)$ term provides noise-level conditioning through a dedicated channel and is likewise shared across tokens within the same denoising sample.
The ReLU clamps effective singular values to non-negative values and lets the router deactivate an expert by pushing its scaling below zero.

Routing adjusts the magnitudes of the rank-one experts but leaves the basis $\vu_d,\vv_d$ fixed.
To allow a small amount of direction-level adaptation, we prepend a product of $L_h$ Householder reflections,
\begin{equation}
    \mH = \prod_{\ell=1}^{L_h} \Big(\mI - 2\,\frac{\vr_\ell\vr_\ell^\top}{\|\vr_\ell\|^2}\Big)
\end{equation}
which is orthogonal by construction~\cite{yuan2024bridging}, so $\mW\mH$ keeps the singular values and the output column space of $\mW$ while rotating the input directions $\vv_d\to\mH^\top\vv_d$ that each expert reads.
The forward pass of a routed layer then reduces to
\begin{equation}
\label{eq:moore_full}
    \vy_{b,q} = \mU\big(\hat{\bm{\sigma}}_{b,q} \odot \mV^\top \mH \vx_{b,q}\big) + \bm{b}
\end{equation}
The reflection vectors $\vr_\ell$ are paired at initialization so that $\mH=\mI$ exactly, and the router reads the unrotated token state $\vx_{b,q}$, which keeps its input distribution aligned with that of the residual and AdaLN paths.

We keep the cross-attention and AdaLN paths shared because they provide visual conditioning and global timestep--state modulation, respectively.
Among the remaining action-token linear maps, we apply full spectral routing only to matrices with well-conditioned spectral coordinates.
For a candidate weight $\mW$, we compute $\kappa_\epsilon(\mW)=\sigma_1/(\sigma_D+\epsilon)$ and route it when $\kappa_\epsilon(\mW)<\tau$; a near-degenerate spectral tail therefore leaves the layer unchanged.
The low-dimensional final action projection is handled separately using task-level spectral offsets only.

\begin{figure}[t]
\centering
\includegraphics[width=\columnwidth]{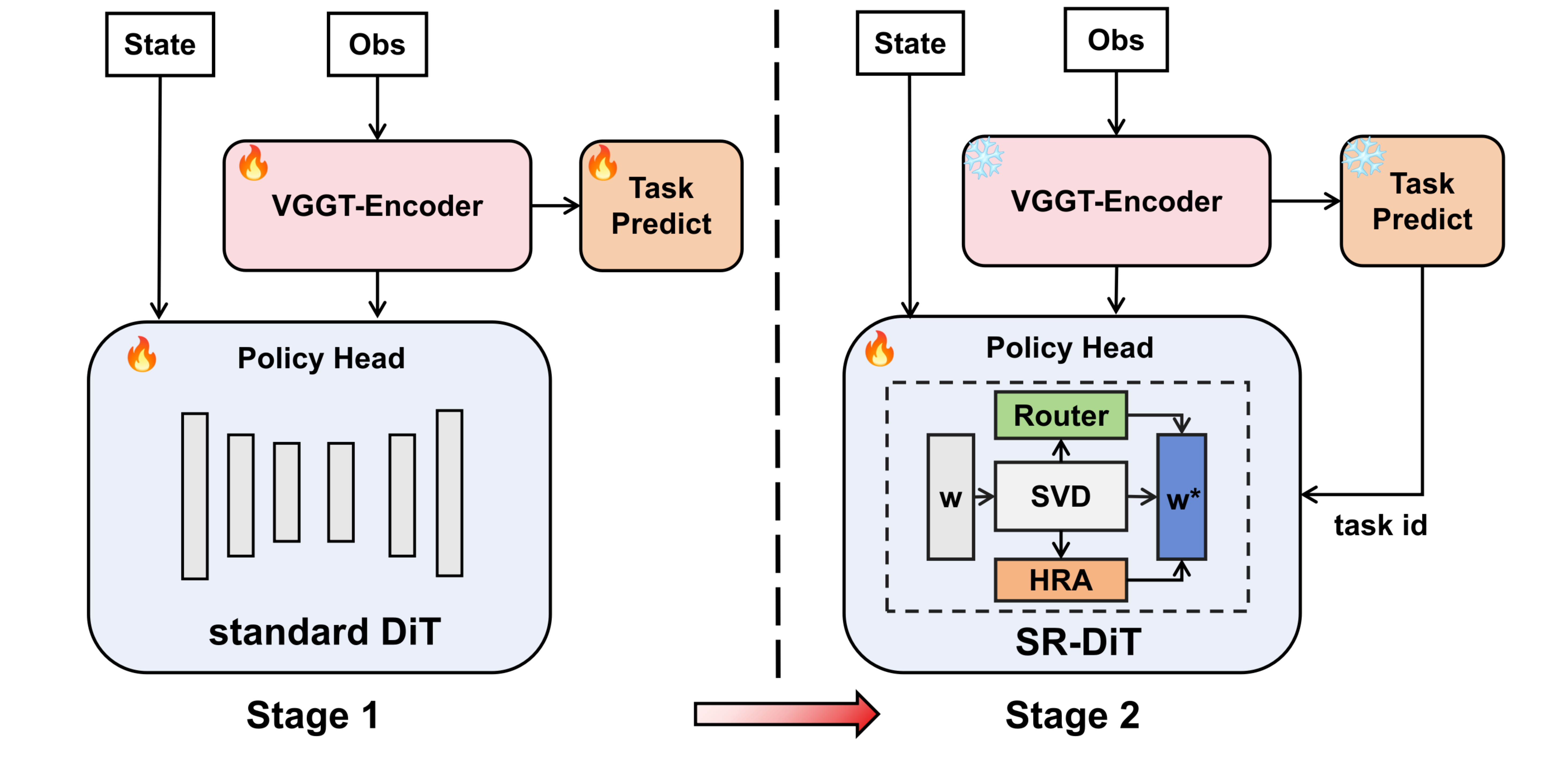}%
\vspace{-6pt}
\caption{\textbf{Two-stage training.} Stage~1 jointly trains the adapter, task head, and a standard DiT denoiser without task conditioning. Stage~2 freezes the visual encoder and task head, decomposes well-conditioned layers by SVD, and updates only the router and Householder vectors.}
\label{fig:two_stage}
\vspace{-10pt}
\end{figure}

%------------------------------------------------------------------------
\subsection{Two-Stage Training}
\label{sec:training}

\method{} is trained in two stages (Fig.~\ref{fig:two_stage}): Stage~1 trains a shared base policy, and Stage~2 converts it into a spectrally routed one.
The separation is required because a weight's SVD basis becomes informative only after the denoiser has converged; decomposing it earlier would route along directions that carry no task structure.

\paragraph{Stage~1}
With the VGGT backbone frozen, we jointly train the spatial-compression adapter, task head, and denoiser on $\bigcup_k\mathcal{D}_k$ under the diffusion loss of Eq.~\ref{eq:ddpm} plus a task-head cross-entropy term,
\begin{equation}
\label{eq:stage1}
    \mathcal{L}_1 = \mathcal{L}_{\text{DM}} + \lambda\,\mathbb{E}_{(\vo_t,k)}\,\mathrm{CE}\big(p(k\mid\vo_t),\,k\big)
\end{equation}
The encoder must produce visual tokens useful for denoising and task recognition, but the denoiser itself receives no task identifier, so Stage~1 learns a shared action backbone while the task head learns the identifier used for Stage~2 routing.

\paragraph{Stage~2}
We freeze the Stage~1 visual encoder and task head, and replace each selected denoiser layer with its rank-one expert form.
The SVD factors $\mU,\mV,\bm{\sigma}$ are frozen; router output layers are zero-initialized so singular-value shifts start at zero, and the Householder reflections use identical non-zero pairs so the converted network exactly reproduces Stage~1 at the start of Stage~2.
We then optimize only the router and Householder parameters under the same diffusion regression as Stage~1, now with the denoiser conditioned on the task identifier $k$,
\begin{equation}
\label{eq:stage2}
    \mathcal{L}_2 = \mathbb{E}_{n,\veps,(\vo_t,\va_t,k)\sim\bigcup_k\mathcal{D}_k} \big\| \veps - \veps_\theta(\va_t^n, n, \vc_t, k) \big\|_2^2
\end{equation}
where $k$ is the ground-truth task label during training and the prediction $\arg\max\mathrm{TaskHead}(\vc^{\text{vis}}_t)$ at inference; all Stage~1 weights stay frozen.
Algorithm~\ref{alg:training} summarizes the procedure.

\begin{algorithm}[t]
    \caption{\method{} training procedure.}
    \label{alg:training}
\begin{algorithmic}
    \STATE \textbf{Stage~1: joint training.}
    \WHILE{not converged}
        \STATE Sample $(\vo, \va, k) \sim \bigcup_k \mathcal{D}_k$;\quad sample $n,\veps$
        \STATE $(\vc^{\text{vis}}, \vc) \leftarrow \mathrm{Encoder}(\vo)$;\quad $\va^n \leftarrow \sqrt{\bar\alpha_n}\va + \sqrt{1-\bar\alpha_n}\veps$
        \STATE $\hat{k} \leftarrow \mathrm{TaskHead}(\vc^{\text{vis}})$
        \STATE $\mathcal{L}_1 \leftarrow \|\veps - \veps_\theta(\va^n, n, \vc)\|^2 + \lambda\,\mathrm{CE}(\hat{k}, k)$
        \STATE Update all trainable parameters on $\nabla \mathcal{L}_1$
    \ENDWHILE
    \STATE
    \STATE \textbf{Stage~2: spectral fine-tuning.}
    \STATE Convert the selected layers to the routed SVD form with zero-shift initialization.
    \WHILE{not converged}
        \STATE Sample $(\vo, \va, k) \sim \bigcup_k \mathcal{D}_k$;\quad sample $n,\veps$
        \STATE $(\vc^{\text{vis}}, \vc) \leftarrow \mathrm{Encoder}(\vo)$;\quad $\va^n \leftarrow \sqrt{\bar\alpha_n}\va + \sqrt{1-\bar\alpha_n}\veps$
        \STATE Update router and Householder params on $\|\veps - \veps_\theta(\va^n, n, \vc, k)\|^2$
    \ENDWHILE
    \STATE At inference, use $k^\star \leftarrow \arg\max \mathrm{TaskHead}(\vc^{\text{vis}})$.
\end{algorithmic}
\end{algorithm}

%========================================================================
\section{Simulation Experiments}
\label{sec:exp}

This section reports six-task RoboTwin results against per-task and multi-task Diffusion Policy baselines ({\S}\ref{sec:main_results}), then analyzes the source of \method{}'s multi-task gains ({\S}\ref{sec:analysis}).

%------------------------------------------------------------------------
\subsection{Experimental Setup}
\label{sec:setup}

The six tasks are drawn from RoboTwin~\cite{mu2024robotwin}: Block Hammer Beat, Bottle Adjust, Container Place, Dual Bottles Pick, Blocks Stack, and Tool Adjust, covering tool use, object reorientation, precision placement, bimanual coordination, sequential assembly, and orientation-aware grasping.
Each task provides 100 teleoperated demonstrations; the robot observes through two RGB cameras (head and front, $240 \times 320$) and 14-DoF joint positions.

We compare against two baseline families.
The \textit{per-task} references train one model per task: DP~\cite{chi2024diffusionpolicy} and VO-DP~\cite{li2025vodp}.
% Setup includes the SDP multi-task baseline.
The \textit{multi-task} baselines train a single model on the union of the six datasets: DP~\cite{chi2024diffusionpolicy} and VO-DP~\cite{li2025vodp} re-trained in the multi-task regime, FDP~\cite{liu2025fdp} which factorizes the action distribution into composable components, and SDP~\cite{wang2024sparse_dp_moe} which uses sparse diffusion experts.
All multi-task entries use the same demonstrations and evaluation protocol.

The visual encoder is a frozen VGGT-1B~\cite{wang2025vggt} ($\ell^\star{=}23$, deterministic resize/center-crop to multiples of $14$), followed by a $14$-token-per-view spatial adapter; the AdaLN-Zero DiT~\cite{peebles2023dit} uses a $100$-step squared-cosine DDPM sampler.
For the selection rule in Sec.~\ref{sec:moore}, we set $\epsilon=10^{-10}$ and $\tau=10^2$.
Applied to the converged Stage~1 denoiser, it retains the fused self-attention QKV projection and two MLP projections in each of the 12 DiT blocks, yielding 36 fully routed matrices; the self-attention output projections remain shared.
Sampling uniformly over concatenated windows, both stages and controls use AdamW with a cosine schedule and evaluate the final checkpoint over $100$ episodes per task across $3$ seeds.
The frozen TaskHead achieves $100\%$ accuracy across all steps of the $1{,}800$ evaluation rollouts.
Multi-task DP, VO-DP, and FDP receive no explicit task ID; SDP, SR-DP Stage~2, Task-AdaLN, and the non-spectral adapter use ground-truth task IDs in training and TaskHead predictions at inference.
Hyperparameters are in Table~\ref{tab:hyperparams}.

\begin{table}[t]
\centering
\caption{Key hyperparameters of \method{}.}
\label{tab:hyperparams}
\vspace{2pt}
\footnotesize
\setlength{\tabcolsep}{2pt}
\begin{tabular}{l c | l c}
\toprule
\textbf{Hyperparam.} & \textbf{Value} & \textbf{Hyperparam.} & \textbf{Value} \\
\midrule
VGGT block $\ell^\star$       & $23$                                  & Task emb.\ $D_t$           & $32$ \\
Pool resolutions              & $\{1{\times}1,2{\times}2,3{\times}3\}$ & Router dim $D_r$           & $32$ \\
Tokens / view                 & $14$                                  & Householder $L_h$          & $4$ \\
DiT blocks $L$                & $12$                                  & Threshold $\tau$           & $10^2$ \\
Embedding dim                 & $384$                                 & DDPM steps                 & $100$ \\
Batch / task wt.            & $200/0.5$                  & Obs./act. horizon    & $1/8$ \\
\midrule
Stage~1 epochs                & $300$                                 & Stage~1 lr                 & $1{\times}10^{-4}$ \\
Stage~2 epochs                & $150$                                 & Stage~2 lr (router/HRA) & $3{\times}10^{-4}$ \\
\bottomrule
\end{tabular}
\vspace{-2pt}
\end{table}

%------------------------------------------------------------------------
\subsection{Main Results}
\label{sec:main_results}

\begin{table*}[t]
\centering
\caption{Success rates (\%) on six RoboTwin tasks. Each entry reports mean$\pm$std over three training seeds, with 100 evaluation episodes per task per seed. Per-task rows are specialist references; \textbf{bold} marks the best single multi-task model in each column.}
\label{tab:main}
\vspace{2pt}
\scriptsize
\renewcommand{\arraystretch}{1.10}
\setlength{\tabcolsep}{2pt}
\begin{tabular*}{\textwidth}{@{\extracolsep{\fill}}l|cccccc@{\hspace{7pt}}|@{\hspace{2pt}}c}
\toprule
\textbf{Method} & \makecell{Block Hammer\\Beat} & \makecell{Bottle\\Adjust} & \makecell{Container\\Place} & \makecell{Dual Bottles\\Pick} & \makecell{Blocks\\Stack} & \makecell{Tool\\Adjust} & \textbf{Avg.\,($\uparrow$)} \\
\midrule
\multicolumn{8}{l}{\textit{Per-task reference (one model per task)}} \\
DP~\cite{chi2024diffusionpolicy}     & 71.0{\scriptsize$\pm$}1.0 & 40.3{\scriptsize$\pm$}2.5 & 15.3{\scriptsize$\pm$}1.2 & 74.7{\scriptsize$\pm$}0.6 & 37.3{\scriptsize$\pm$}1.2 & 34.0{\scriptsize$\pm$}1.0 & 45.4{\scriptsize$\pm$}1.0 \\
VO-DP~\cite{li2025vodp}              & 83.7{\scriptsize$\pm$}1.5 & 64.0{\scriptsize$\pm$}4.6 & 43.0{\scriptsize$\pm$}3.0 & 85.7{\scriptsize$\pm$}1.5 & 54.3{\scriptsize$\pm$}3.2 & 57.0{\scriptsize$\pm$}1.7 & 64.6{\scriptsize$\pm$}0.6 \\
\midrule
\multicolumn{8}{l}{\textit{Multi-task (single model on the six tasks)}} \\
DP~\cite{chi2024diffusionpolicy} (multi-task) & 40.0{\scriptsize$\pm$}1.7 & 14.3{\scriptsize$\pm$}2.5 & 14.3{\scriptsize$\pm$}4.0 & 62.3{\scriptsize$\pm$}0.6 & 23.0{\scriptsize$\pm$}1.0 & 25.7{\scriptsize$\pm$}1.5 & 29.9{\scriptsize$\pm$}0.5 \\
VO-DP~\cite{li2025vodp} (multi-task) & 42.0{\scriptsize$\pm$}2.6 & 16.0{\scriptsize$\pm$}4.0 & 40.7{\scriptsize$\pm$}1.2 & 91.0{\scriptsize$\pm$}1.7 & 32.0{\scriptsize$\pm$}2.6 & 58.0{\scriptsize$\pm$}0.0 & 46.6{\scriptsize$\pm$}0.3 \\
FDP~\cite{liu2025fdp}                & 62.3{\scriptsize$\pm$}1.5 & 29.7{\scriptsize$\pm$}1.5 & 35.7{\scriptsize$\pm$}3.1 & 52.7{\scriptsize$\pm$}1.2 & 24.0{\scriptsize$\pm$}3.5 & 32.3{\scriptsize$\pm$}4.0 & 39.4{\scriptsize$\pm$}1.6 \\
% SDP multi-task baseline uses the three-training-seed statistics defined in the caption.
SDP~\cite{wang2024sparse_dp_moe} & 64.3{\scriptsize$\pm$}2.5 & 24.0{\scriptsize$\pm$}3.6 & 37.0{\scriptsize$\pm$}4.0 & 69.7{\scriptsize$\pm$}2.1 & 32.0{\scriptsize$\pm$}4.0 & 40.0{\scriptsize$\pm$}4.0 & 44.5{\scriptsize$\pm$}1.6 \\
\midrule
\method{} (Stage~1)                  & 90.0{\scriptsize$\pm$}0.0 & 41.3{\scriptsize$\pm$}1.2 & 33.3{\scriptsize$\pm$}2.9 & 90.7{\scriptsize$\pm$}0.6 & 30.0{\scriptsize$\pm$}1.7 & 50.3{\scriptsize$\pm$}0.6 & 55.9{\scriptsize$\pm$}0.8 \\
\method{} (Stage~2)                  & \best{95.0{\scriptsize$\pm$}1.0} & \best{53.3{\scriptsize$\pm$}0.6} & \best{53.7{\scriptsize$\pm$}3.5} & \best{91.3{\scriptsize$\pm$}1.5} & \best{53.3{\scriptsize$\pm$}3.1} & \best{63.7{\scriptsize$\pm$}3.2} & \best{68.4{\scriptsize$\pm$}0.8} \\
\bottomrule
\end{tabular*}
\vspace{-10pt}
\end{table*}

Without a vision-foundation encoder, per-task DP averages $45.4\%$ across six tasks (Table~\ref{tab:main}), dropping to $29.9\%$ under joint training.
VO-DP reaches $64.6\%$ with per-task training but drops to $46.6\%$ when trained jointly across all six tasks.
FDP, the action-distribution factorization baseline, reaches $39.4\%$ under joint training, suggesting that factorization alone is insufficient to close the observed performance gap.
\method{} reaches $55.9\%$ in Stage~1 and $68.4\%$ in Stage~2, surpassing per-task VO-DP by $3.8$ points, multi-task VO-DP by $21.8$, and FDP by $29.0$.
Stage~1 matches or exceeds per-task VO-DP on Block Hammer Beat and Dual Bottles Pick, where shared action structure is largely sufficient, but trails on the other four tasks; Stage~2 closes most of this gap and surpasses per-task VO-DP on Container Place and Tool Adjust.
The analysis below shows these per-task gains coincide with less opposition between task-wise spectral-coordinate loss sensitivities.

%------------------------------------------------------------------------
\subsection{Mechanism Analysis}
\label{sec:analysis}

% Training-duration and task-conditioning controls are reported with the three component ablations.
\begin{table}[t]
\centering
\caption{Simulation ablations (mean$\pm$std success, \%).}
\label{tab:ablation}
\scriptsize
\setlength{\tabcolsep}{2pt}
\begin{tabular}{lcccccc}
\toprule
Setting & \makecell{Task\\cond.} & \makecell{Spectral\\route} & HRA & \makecell{Router\\time} & \makecell{Layer\\select.} & Avg. \\
\midrule
Continued Stage~1 & $\times$ & $\times$ & $\times$ & $\times$ & $\times$ & 56.1$\pm$0.3 \\
Task-AdaLN & $\checkmark$ & $\times$ & $\times$ & $\times$ & $\times$ & 56.1$\pm$0.4 \\
Non-spectral adapter & $\checkmark$ & $\times$ & $\times$ & $\checkmark$ & $\checkmark$ & 63.3$\pm$0.9 \\
w/o router timestep & $\checkmark$ & $\checkmark$ & $\checkmark$ & $\times$ & $\checkmark$ & 63.8$\pm$0.7 \\
w/o Householder & $\checkmark$ & $\checkmark$ & $\times$ & $\checkmark$ & $\checkmark$ & 60.9$\pm$0.3 \\
Expanded routing & $\checkmark$ & $\checkmark$ & $\checkmark$ & $\checkmark$ & $\times$ & 65.5$\pm$0.3 \\
\rowcolor{gray!20}
Full SR-DP & $\checkmark$ & $\checkmark$ & $\checkmark$ & $\checkmark$ & $\checkmark$ & \best{68.4$\pm$0.8} \\
\bottomrule
\end{tabular}
\vspace{-6pt}
\end{table}

% ---- Fig: Encoder feature t-SNE ----
\begin{figure}[t]
\centering
\includegraphics[width=\columnwidth]{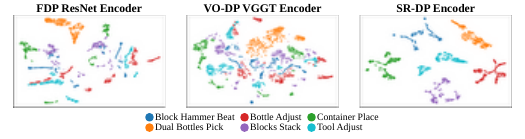}%
\vspace{-10pt}
\caption{\textbf{Encoder feature t-SNE.} Visual conditions from the complete FDP, multi-task VO-DP, and \method{} pipelines over $9{,}600$ held-out frames ($1{,}600$/task) across the six RoboTwin tasks. FDP uses ResNet, VO-DP pools VGGT without task loss, and \method{} pairs VGGT with the spatial adapter and task head.}
\label{fig:encoder_tsne}
\vspace{-10pt}
\end{figure}

% ---- Fig: Gradient interference in SVD coordinates ----
\begin{figure}[t]
\centering
\includegraphics[width=\columnwidth]{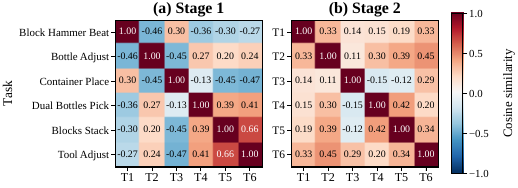}%
\vspace{-6pt}
\caption{\textbf{Loss-sensitivity conflict in the shared SVD basis.} Layer-averaged task-pair sensitivity cosine ($4$ held-out windows/task, $n{=}50$, shared noise; negative: opposing sensitivities). Stage~2 lowers the mean conflict score from $0.267$ to $0.104$.}
\label{fig:gradient_interference}
\vspace{-10pt}
\end{figure}

% ---- Fig: Router expert activation along demonstrations ----
\begin{figure}[t]
\centering
\includegraphics[width=\columnwidth]{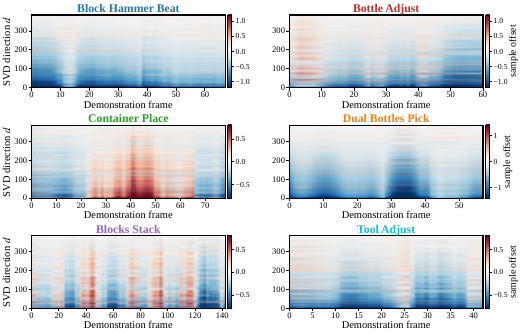}%
\vspace{-6pt}
\caption{\textbf{Router activation along held-out demonstrations.} Six RoboTwin tasks for the routed layer with maximum temporal std (self-attention QKV in Block~9), scanned every third frame of one held-out episode/task ($n{=}50$). Color shows token-state offset averaged over action tokens.}
\label{fig:router_rollout_experts}
\vspace{-10pt}
\end{figure}

% Component ablations are discussed using quantitative results and observed execution behavior.
\paragraph{Component ablations}
All variants in Table~\ref{tab:ablation} share the data, Stage~1 checkpoint, and $150$ cosine-scheduled AdamW epochs ($1{\times}10^{-4}$ for Continued Stage~1; $3{\times}10^{-4}$ for Stage~2 adapters).
Task-AdaLN ($0.45$M; task embedding and zero-initialized AdaLN) matches Continued Stage~1 at $56.1\%$ vs.\ $68.4\%$ for SR-DP ($1.2$M).
A same-layer non-spectral adapter ($1.2$M; task, token state, and step on the $36$ QKV/MLP maps without SVD or Householder) reaches $63.3\%$ and regresses on one task.
Removing router timestep or Householder drops $4.6$ and $7.5$ points, and expanded routing ($60$ maps, adding self-/cross-attention outputs) drops $2.9$ points despite more parameters.

\paragraph{Visual representation separability}
Fig.~\ref{fig:encoder_tsne} compares the complete conditioning pipelines over dense frames from the six RoboTwin tasks.
FDP uses a ResNet encoder, VO-DP pools VGGT features without task loss, and \method{} pairs the spatial-compression adapter with the auxiliary task-prediction head.
VO-DP's conditions remain mixed across tasks, whereas \method{} forms compact task-specific regions, indicating that pairing the spatial adapter with the task loss yields a more discriminative condition.

\paragraph{Spectral-coordinate sensitivity conflict}
Sensitivity conflict is measured in rank-one coordinates indexed by the Stage~1 SVD (Fig.~\ref{fig:gradient_interference}).
Stage~1 uses projected dense-weight gradients; Stage~2 probes loss sensitivities to the frozen singular values along the corresponding Householder-rotated directions.
For routed layer $\ell$, define $c_{ij}^\ell=\cos(\bm{g}_i^\ell,\bm{g}_j^\ell)$ and $D^\ell=\frac{1}{K(K-1)}\sum_{i\ne j}\max(0,-c_{ij}^\ell)$.
We report $D=\frac{1}{36}\sum_{\ell=1}^{36}D^\ell$, taking the negative part before layer averaging; the heatmap instead shows $\bar{c}_{ij}=\frac{1}{36}\sum_{\ell=1}^{36}c_{ij}^\ell$.
The score drops from $0.267$ to $0.104$, indicating less opposition among spectral-coordinate loss sensitivities.

% ---- Fig: Simulation qualitative rollouts ----
\begin{figure}[t]
\centering
\scalebox{1}[0.968]{\includegraphics[width=0.95\columnwidth,trim=0 35bp 0 0,clip]{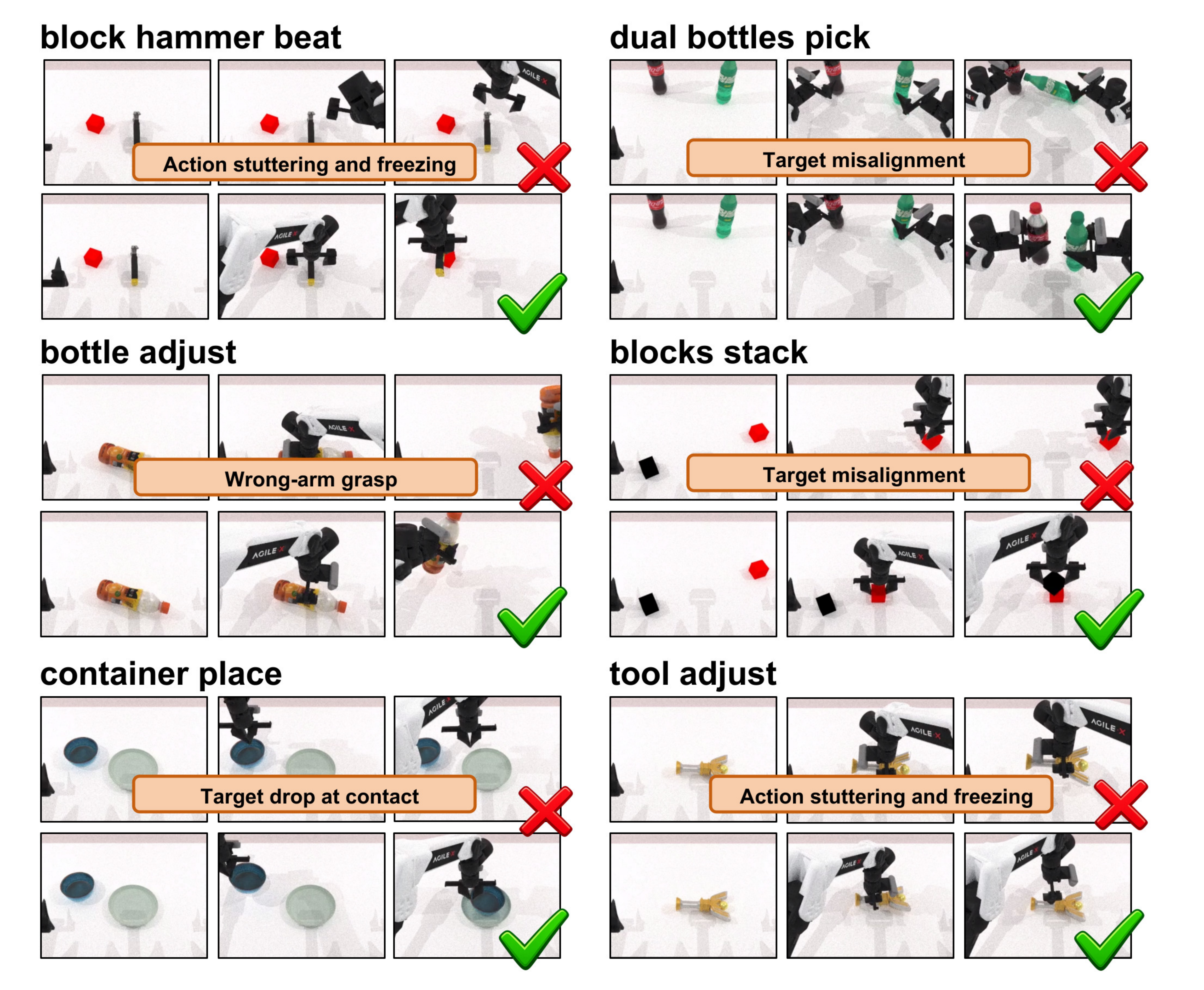}}%
\vspace{-7pt}
\caption{Qualitative comparison on the six RoboTwin tasks: each task pairs a selected Stage~1 failure (top) and Stage~2 success (bottom) under the same initial conditions to illustrate failure modes.}
\label{fig:sim_qualitative}
\vspace{-7pt}
\end{figure}

\paragraph{Dynamic routed specialization}
Fig.~\ref{fig:router_rollout_experts} visualizes the token-state router along held-out demonstration trajectories.
Rows denote SVD rank-one expert directions and columns denote demonstration frames; color is the token-state offset averaged over action tokens before adding the task offset and base singular value.
Active directions vary over time and across tasks, indicating phase-dependent activation rather than a static task-level correction.
The qualitative rollouts in Fig.~\ref{fig:sim_qualitative} show the corresponding action-level effect.
In these examples, baseline policies produce the correct action plan, but execution breaks down at fine-manipulation moments---action stuttering and freezing, wrong-arm grasps, inaccurate target alignment, and target drops at the contact point.
\method{} leaves the overall action trajectory essentially unchanged but stabilises these details, with smoother continuation, tighter alignment, and more reliable contact.
In these rollouts, the improvement concentrates in the final action segment, consistent with \method{}'s design: task-specific routing within the frozen spectral basis rather than rewriting the action plan.

Policy-head trainable parameters (excluding VGGT) are $28.4$M (Stage~1), $31.2$M (FDP), $0.45$M (Task-AdaLN), and $1.2$M for both the non-spectral adapter and Stage~2 ($4.1\%$ of $29.6$M).

%========================================================================
\section{Real-World Experiments}
\label{sec:real}

The real-world evaluation tests whether the Stage~2 routing gains observed in simulation transfer to physical bimanual manipulation with shared objects and workspace.

\subsection{Platform and Tasks}
\label{sec:real_platform}

% ---- Fig: Real-world hardware setup ----
\begin{figure}[t]
\centering
\scalebox{1}[0.968]{\includegraphics[width=0.86\columnwidth]{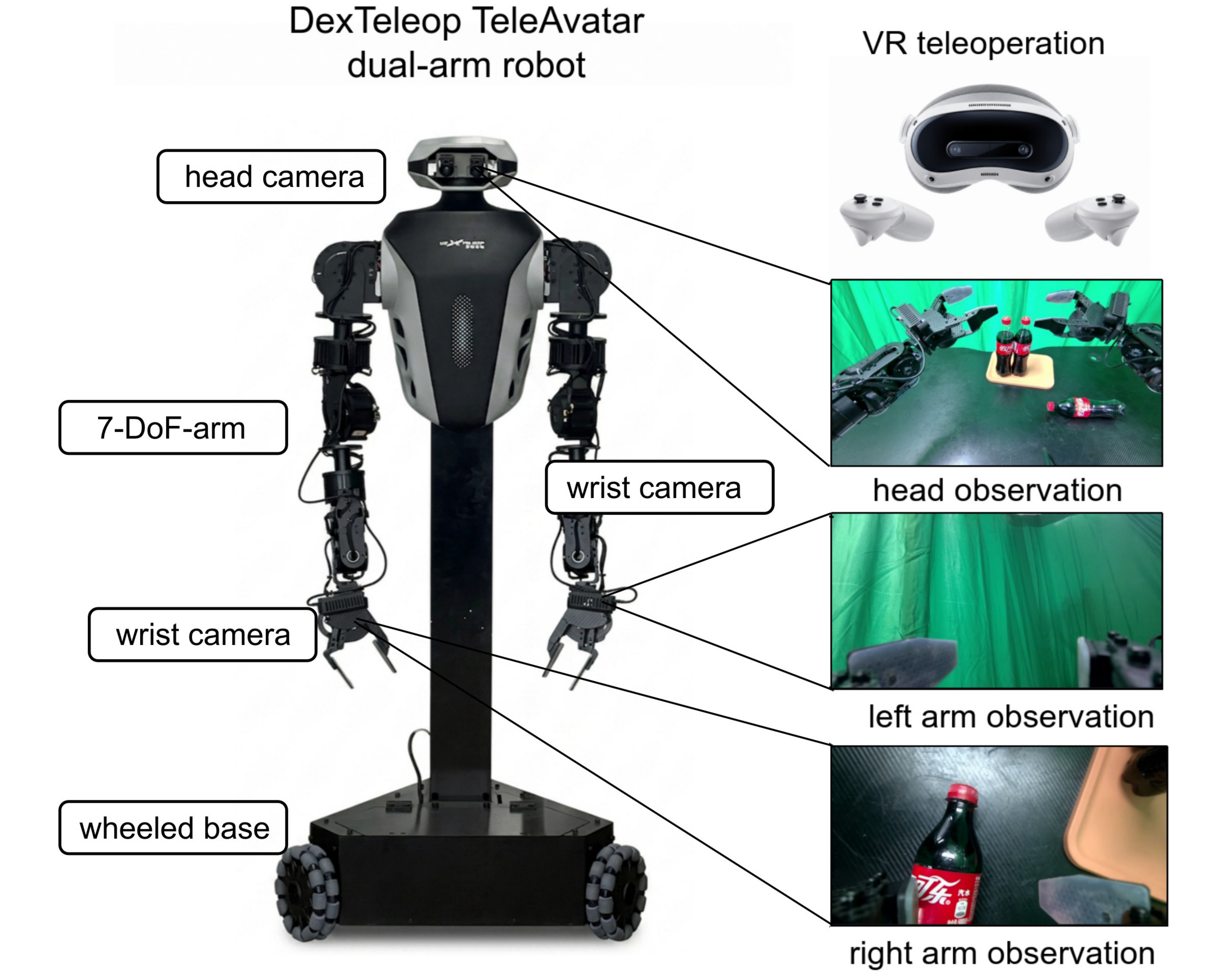}}%
\vspace{-4pt}
\caption{\textbf{Real-world platform.} The DexTeleop TeleAvatar dual-arm robot, teleoperated through a VR headset and controllers for demonstration collection. Three RGB cameras provide observations: a head-mounted view and two wrist-mounted views (one per arm).}
\label{fig:real_setup}
\vspace{-10pt}
\end{figure}

The platform is the TeleAvatar (TA lite) dual-arm system~\cite{dexteleop} (Fig.~\ref{fig:real_setup}), equipped with two 7-DoF arms on a wheeled base and parallel-jaw grippers.
The workspace is a $77$\,cm table with a staging tray on the front-right and a reject bin on the front-left, both reachable by either arm.
Three RGB cameras (one head-mounted and one on each wrist) provide $240\times320$ images.
VR teleoperation records expert demonstrations at $30$\,Hz with synchronized RGB streams, 14-DoF joint positions, and gripper state.

The four tasks share the same robot, workspace, fixed fixtures, and geometrically similar $500$\,mL bottles.
Task-relevant differences include bottle identity, position, orientation, and organizer occupancy, requiring the policy to distinguish different action requirements from local visual cues within similar scenes.
For each task, we collect $200$ demonstrations with object positions and orientations randomized across episodes, yielding four independent datasets.

T1 (\emph{Sort out the Sprite}) presents a three-cell organizer containing one Sprite bottle and two Coke bottles with randomized cell indices; the robot must remove the Sprite and drop it in the reject bin, discriminating the target through label color.
T2 (\emph{Upright the bottle}) requires the robot to pick a fallen Coke bottle lying in arbitrary orientation on the table, reorient it vertically, and place it in a free slot of the staging tray.
T3 (\emph{Flip the inverted Coke}) places one cap-down Coke bottle and two upright Coke bottles in the organizer; the robot must grasp the inverted bottle, flip it $180^\circ$ in mid-air, and return it to its original cell without intermediate regrasps.
T4 (\emph{Refill the organizer}) starts with $0$, $1$, or $2$ upright Coke bottles in the organizer and requires the robot to transfer bottles from the staging tray until all three cells are filled, producing a variable-length action sequence.

% ---- Fig: Real-world qualitative rollouts ----
\begin{figure*}[t]
\centering
\includegraphics[width=0.94\textwidth]{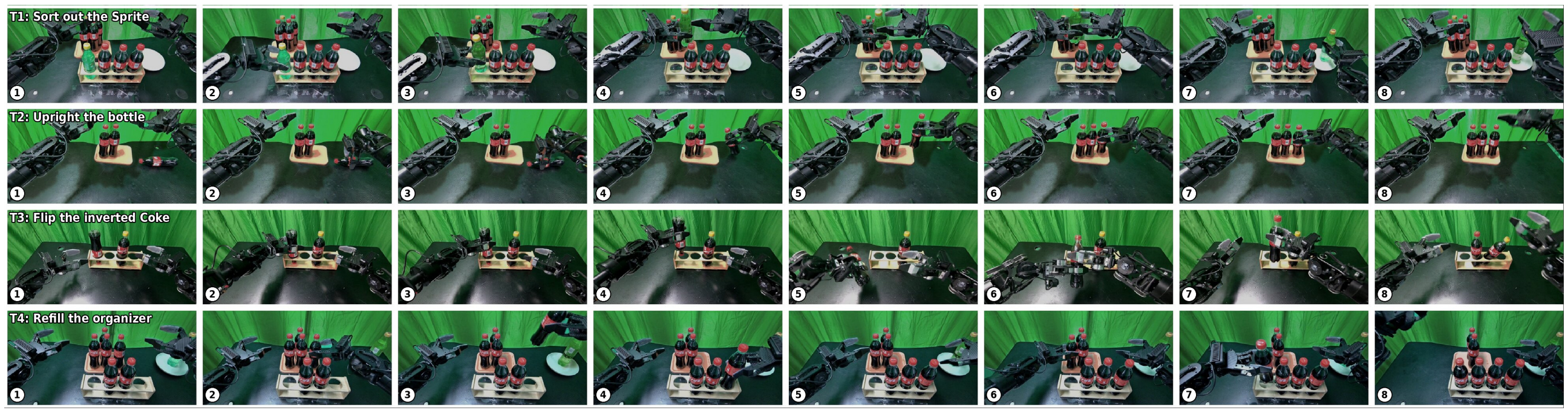}%
\vspace{-4pt}
\caption{Real-world rollouts of \method{} on the four TeleAvatar tasks. Rows correspond to T1--T4, and columns show temporally ordered key frames from one successful Stage~2 rollout per task.}
\label{fig:real_qualitative}
\vspace{-10pt}
\end{figure*}

\subsection{Results}
\label{sec:real_results}

We train one \method{} model jointly on the union of the four datasets.
Stage~1 and Stage~2 train for $150$ and $80$ epochs, respectively, with one-frame observations and a $30$-step action horizon.
Each task is evaluated over $50$ rollouts from the same predefined robot home configuration. Object poses and task-relevant scene states are independently randomized for Stage~1 and Stage~2 within the same feasible ranges.
A rollout succeeds when the task's post-condition holds and both arms return to the home pose without misplacing any object.
On the $200$ Stage~2 initial rollout observations ($50$/task), the frozen TaskHead achieves $99.0\%$ ($198/200$) accuracy.
We compare Stage~1, Task-AdaLN ($0.45$M, same $3{\times}10^{-4}$ AdamW schedule), and SR-DP Stage~2 ($1.2$M) under identical data, hardware, and protocol; Task-AdaLN tests task-conditioned modulation without spectral routing.

Stage~2 increases average success from $24.0\%$ to $43.0\%$ ($+19.0$ percentage points; Table~\ref{tab:real}, Fig.~\ref{fig:real_qualitative}).
Task-AdaLN reaches $23.5\%$, $0.5$ points below Stage~1 and $19.5$ points below SR-DP Stage~2.
The largest gains over Stage~1 are on T4 ($+26$ points; variable-length refilling) and T1 ($+24$ points; target discrimination among visually similar bottles).
Because T3 shifts only from $13/50$ to $15/50$ under unpaired scenes, we interpret the gain at the aggregate level.

\begin{table}[t]
\centering
\caption{Real-world success counts over $50$ rollouts per task. The final column reports the average success rate (\%) across four tasks.}
\label{tab:real}
\vspace{2pt}
\setlength{\tabcolsep}{3pt}
\begin{tabular}{lccccc}
\toprule
Method & T1 & T2 & T3 & T4 & Avg. (\%) \\
\midrule
\method{} (Stage~1) & 7/50 & 12/50 & 13/50 & 16/50 & 24.0 \\
% Task-AdaLN successes on T1--T4: 12, 8, 11, and 16, totaling 47/200.
Task-AdaLN & 12/50 & 8/50 & 11/50 & 16/50 & 23.5 \\
\rowcolor{gray!20}
\method{} (Stage~2) & \best{19/50} & \best{23/50} & \best{15/50} & \best{29/50} & \best{43.0} \\
\bottomrule
\end{tabular}
\vspace{-10pt}
\end{table}

%========================================================================
\section{Conclusion}
\label{sec:conclusion}

We presented \method{}, a multi-task diffusion policy addressing the per-task accuracy drop under joint training across manipulation tasks.
\method{} combines task-discriminative visual conditioning with compact spectral adaptation of a shared diffusion denoiser.
Because the experts share a frozen orthogonal basis, task-specific adaptation uses spectral coefficients and Householder input rotations, limiting direct updates to shared denoiser weights.
Experiments show that a single \method{} model surpasses per-task baselines on average on RoboTwin and improves aggregate success over Stage~1 and Task-AdaLN on TeleAvatar; Stage~2 trains $4.1\%$ of policy-head parameters.

The SVD basis is meaningful only after Stage~1 has converged, so the procedure is not end-to-end.
Language-conditioned routing and a single-stage training objective are natural next steps.

% ============================================================
%  Bibliography
% ============================================================
\bibliographystyle{IEEEtran}
\bibliography{reference}

\end{document}